\pdfoutput=1

\documentclass[11pt]{article}

\usepackage{emnlp2021}

\usepackage{times}
\usepackage{latexsym}

\usepackage{booktabs}
\usepackage{multirow}

\usepackage[T1]{fontenc}

\usepackage[utf8]{inputenc}

\usepackage{microtype}

\usepackage[draft]{todonotes}   
\usepackage{xcolor}
\usepackage{lipsum} 

\title{Avoiding Inference Heuristics in Few-shot Prompt-based Finetuning}

\author{Prasetya Ajie Utama$^{\dag\ddag}$ , Nafise Sadat Moosavi$^{\ddag}$, Victor Sanh$^{\clubsuit}$, Iryna Gurevych$^{\ddag}$\\
  \\
  $^{\dag}$Research Training Group AIPHES\\
  $^{\ddag}$UKP Lab, Technische Universität Darmstadt\\
  $^{\clubsuit}$Hugging Face, Brooklyn, USA \\
  $^{\ddag}$\url{https://www.ukp.tu-darmstadt.de}\\
  {\small\url{utama@ukp.tu-darmstadt.de}}
  }

\begin{document}
\maketitle
\begin{abstract}
Recent \textit{prompt-based} approaches allow pretrained language models to achieve strong performances on \textit{few-shot finetuning} by reformulating downstream tasks as a language modeling problem. In this work, we demonstrate that, despite its advantages on low data regimes, finetuned prompt-based models for sentence pair classification tasks still suffer from a common pitfall of adopting inference heuristics based on lexical overlap, e.g., models incorrectly assuming a sentence pair is of the same meaning because they consist of the same set of words. Interestingly, we find that this particular inference heuristic is significantly less present in the zero-shot evaluation of the prompt-based model, indicating how finetuning can be \textit{destructive} to useful knowledge learned during the pretraining. We then show that adding a regularization that preserves pretraining weights is effective in mitigating this destructive tendency of few-shot finetuning. Our evaluation on three datasets demonstrates promising improvements on the three corresponding challenge datasets used to diagnose the inference heuristics.\footnote{The code is available at \url{https://github.com/UKPLab/emnlp2021-prompt-ft-heuristics}}


\end{abstract}

\section{Introduction}
Prompt-based finetuning has emerged as a promising paradigm to adapt Pretrained Language Models (PLM) for downstream tasks with limited number of labeled examples \cite{schick-schutze-2021-exploiting, radford2019language}. This approach reformulates downstream task instances as a language modeling input,\footnote{E.g., appending a cloze prompt ``It was \texttt{[MASK]}'' to a sentiment prediction input sentence ``Delicious food!'', and obtaining the sentiment label by comparing the probabilities assigned to the words ``great'' and ``terrible''.} allowing PLMs to make non-trivial task-specific predictions even in zero-shot settings. This in turn, provides a good initialization point for data efficient finetuning \cite{gao-etal-2021-making}, resulting in a strong advantage on low data regimes where the standard finetuning paradigm struggles. However, the success of this prompting approach has only been shown using common held-out evaluations, which often conceal certain undesirable behaviors of models \cite{niven-kao-2019-probing}.

One such behavior commonly reported in downstream models is characterized by their preference to use surface features over general linguistic information \cite{warstadt-etal-2020-learning}. In the Natural Language Inference (NLI) task, \citet{mccoy-etal-2019-right} documented that models preferentially use the lexical overlap feature between sentence pairs to blindly predict that one sentence \textit{entails} the other.
Despite models' high in-distribution performance,
they often fail on counterexamples of this  \textit{inference heuristic}, e.g., they predict that ``\textit{the cat chased the mouse}'' entails ``\textit{the mouse chased the cat}''.


At the same time, there is a mounting evidence that pre-training on large text corpora extracts rich linguistic information \cite{hewitt-manning-2019-structural, tenney-etal-2019-bert}. However, based on recent studies, standard finetuned models often overlook this information in the presence of lexical overlap \cite{Nie_Wang_Bansal_2019, dasgupta2018evaluating}. 
We therefore question whether direct adaptation of PLMs using prompts can better transfer the use of this information during finetuning. We investigate this question by systematically studying the heuristics in a prompt-based model finetuned across three datasets with varying data regimes. Our intriguing results reveal that: (i) zero-shot prompt-based models are more robust to using the lexical overlap heuristic during inference, indicated by their high performance on the corresponding challenge datasets; (ii) however, prompt-based finetuned models quickly adopt this heuristic as they learn from more labeled data, which is indicated by gradual degradation of the performance in challenge datasets.

We then show that regularizing prompt-based finetuning, by penalizing the learning from updating the weights too far from their original pretrained values, is an effective approach to improve the in-distribution performance on target datasets, while mitigating the adoption of inference heuristics. 
Overall, our work suggests that while prompt-based finetuning has gained impressive results on standard benchmarks, it can has a negative impact regarding inference heuristics, which in turn suggests the importance of a more thorough evaluation setup to ensure meaningful progress.


\section{Inference Heuristics in Prompt-based Finetuning}
\paragraph{Prompt-based PLM Finetuning}
In this work, we focus on sentence pairs classification tasks, where the goal is to predict semantic relation $y$ of an input pair $x=(s_1,s_2)$. In a standard finetuning setting, $s_1$ and $s_2$ are concatenated along with a special token \texttt{[CLS]}, whose embedding is used as an input to a newly initialized \textit{classifier head}.

The \textit{prompt-based} approach, on the other hand, reformulates pair $x$ as a masked language model input using a pre-defined template and word-to-label mapping. For instance, \citet{schick-schutze-2021-exploiting} formulate a natural language inference instance $(s_1,s_2,y)$ as: $$\texttt{[CLS]} s_1 \texttt{?} \texttt{[MASK]} \textrm{, } s_2 \texttt{[SEP]}$$
with the following mapping for the masked token: ``yes''$\rightarrow$ ``entailment'', ``maybe''$\rightarrow$``neutral'', and ``no'' $\rightarrow$ ``contradiction''. The probabilities assigned by the PLM to the label words at the \texttt{[MASK]} token can then be directly used to make task-specific predictions, allowing PLM to perform in a zero-shot setting. Following \citet{gao-etal-2021-making}, we further finetune the prompt-based model on the available labeled examples for each task. Note that this procedure finetunes only the existing pre-trained weights, and does not introduce new parameters.



\begin{table}
\footnotesize
\centering
 \begin{tabular}{lp{0.33\textwidth}} 
 \toprule
 \multicolumn{2}{c}{\textbf{Original Input}}\\
 \midrule
 \textbf{Premise} & The actors that danced saw the author. \\
 \textbf{Hypothesis} & The actors saw the author. \\
 \textbf{Label} & entailment \textbf{(support)}\\
 \midrule
 \textbf{Premise} & The managers near the scientist resigned. \\
 \textbf{Hypothesis} & The scientist resigned. \\
 \textbf{Label} & non-entailment \textbf{(against)}\\
 \midrule
 \multicolumn{2}{c}{\textbf{Reformulated Input}}\\
 \midrule
 \textbf{Premise} & The actors that danced saw the author? \texttt{[MASK],} the actors saw the author. \\
 \textbf{Label word} & \textit{Yes} \\
 \midrule
 \textbf{Premise} & The managers near the scientist resigned? \texttt{[MASK],} the scientist resigned. \\
 \textbf{Label word} & \textit{No} / \textit{Maybe} \\
 \bottomrule
 \end{tabular}
 \caption{\textbf{Top:} input examples of the NLI task that \textbf{support} or are \textbf{against} the lexical overlap heuristics. \textbf{Bottom:} reformulated NLI instances as masked language model inputs with the expected label words.}
 \label{tab:illustration}
\end{table}

\paragraph{Task and Datasets}
We evaluate on three English language datasets included in the GLUE benchmark \cite{wang2018glue} for which there are challenge datasets to evaluate the lexical overlap heuristic: MNLI \cite{williams-etal-2018-broad}, SNLI \cite{bowman-etal-2015-large}, and Quora Question Pairs (QQP). In MNLI and SNLI, the task is to determine whether premise sentence $s_1$ \textit{entails}, \textit{contradicts}, or is \textit{neutral} to the hypothesis sentence $s_2$. In QQP, $s_1$ and $s_2$ are a pair of questions that are labeled as either \textit{duplicate} or \textit{non-duplicate}.

Researchers constructed corresponding \textit{challenge} sets for the above datasets, which are designed to contain examples that are \textit{against} the heuristics, i.e., the examples exhibit word overlap between the two input sentences but are labeled as non-entailment for NLI or non-duplicate for QQP. We evaluate each few-shot model against its corresponding challenge dataset. Namely, we evaluate models trained on MNLI against entailment and non-entailment subsets of the HANS dataset \cite{mccoy-etal-2019-right}, which are further categorized into lexical overlap (lex.), subsequence (subseq.), and constituent (const.) subsets; SNLI models against the long and short subsets of the Scramble Test challenge set \cite{dasgupta2018evaluating}; and QQP models against the PAWS dataset \cite{zhang-etal-2019-paws}.\footnote{See appendix \ref{implementation} for details of HANS, PAWS, and Scramble Test test sets.} We illustrate challenge datasets examples and their reformulation as prompts in Table \ref{tab:illustration}.


\begin{figure*}%
\centering
\includegraphics[height=3.2cm]{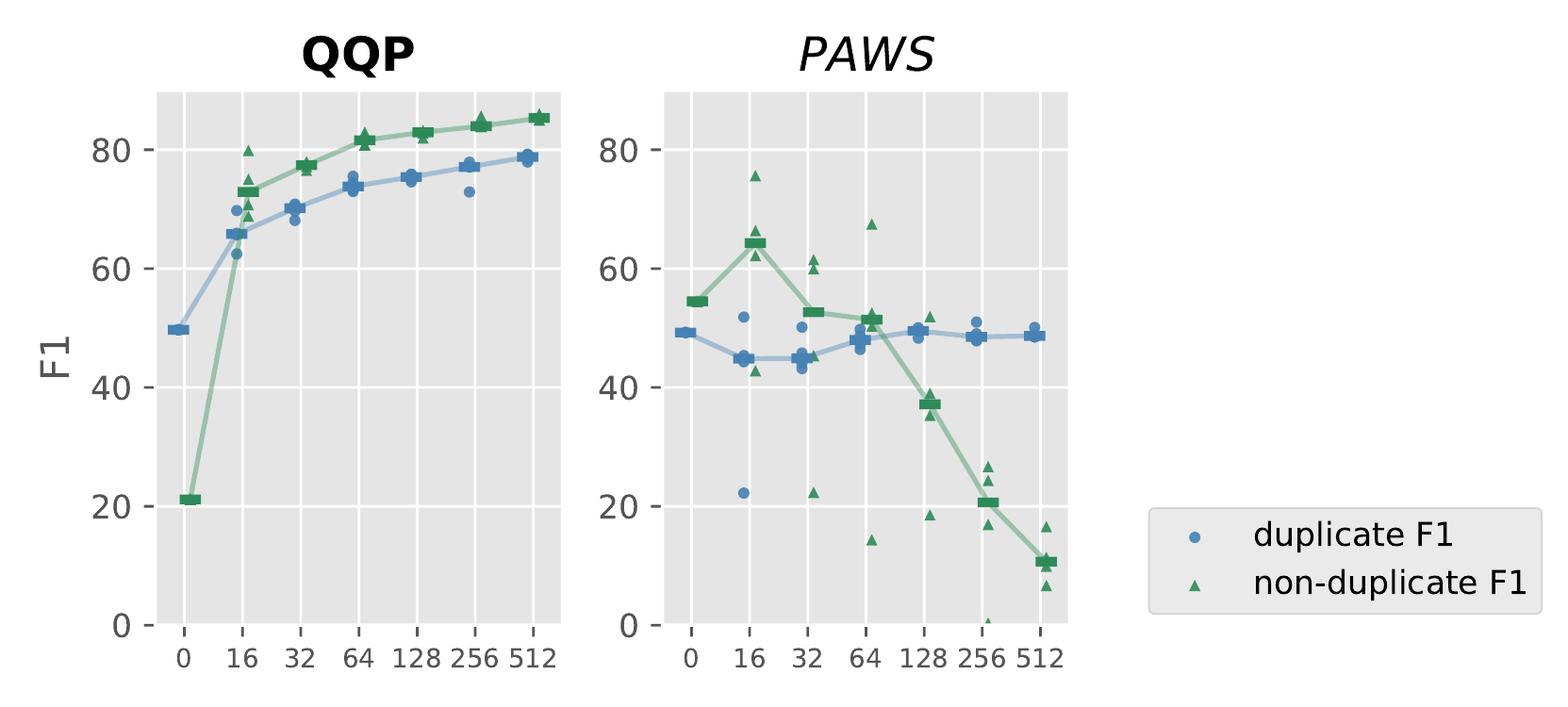}
\includegraphics[height=3.2cm]{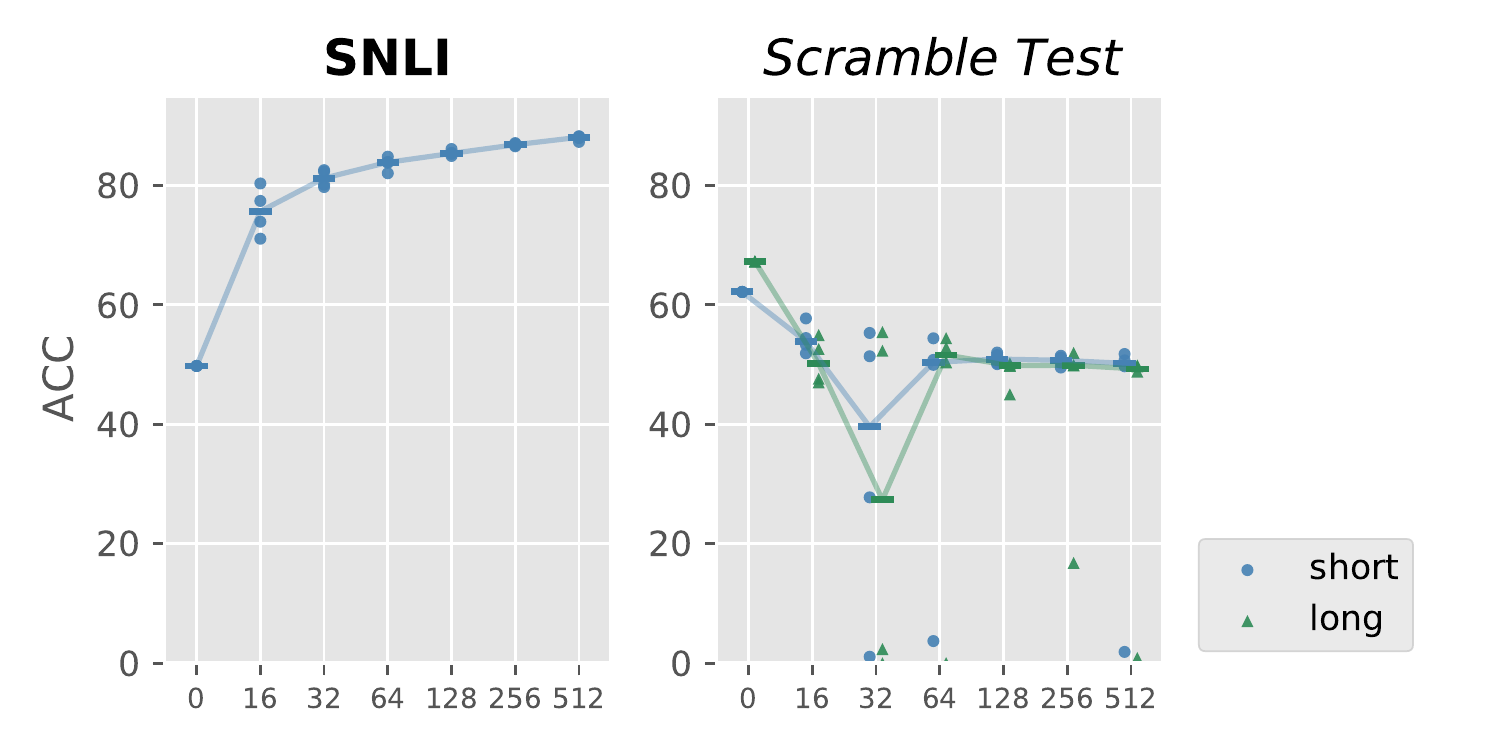}
\includegraphics[width=.7\linewidth]{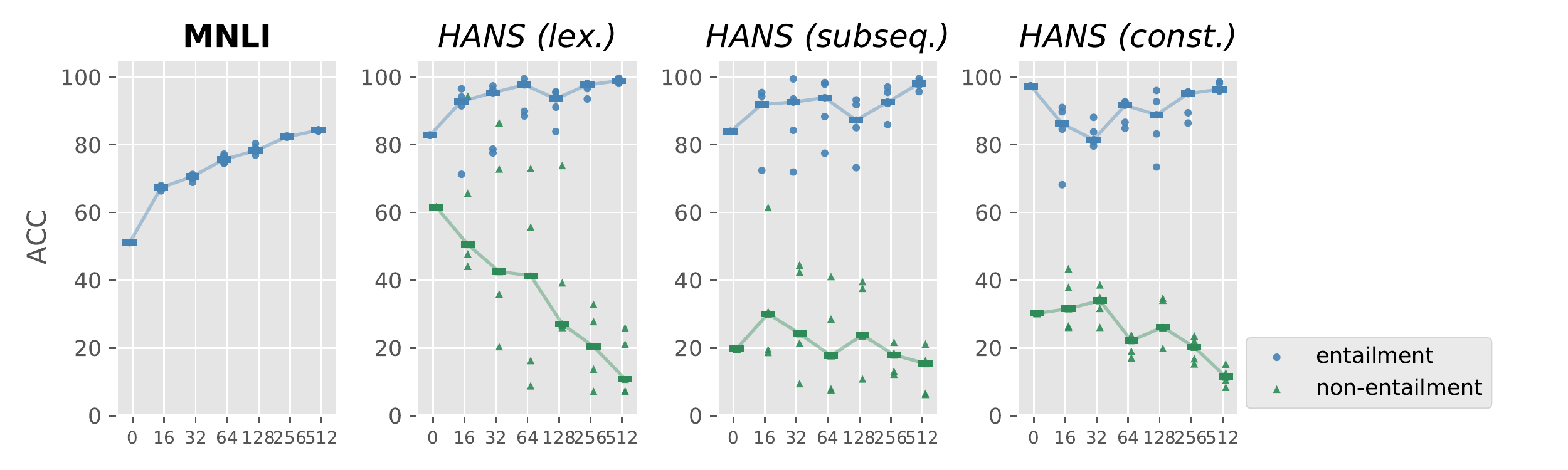}
\caption{In-distribution (\textbf{bold}) vs. challenge datasets (\textit{italic}) evaluation results of prompt-based finetuning across different data size $\mathbf{K}$ (x axis), where $\mathbf{K}=0$ indicates zero-shot evaluation. In all challenge sets, the overall zero-shot performance (both blue and green plots) degrades as the model is finetuned using more data. 
}
\label{fig:data_regimes}
\end{figure*}

\paragraph{Model and Finetuning}
Our training and standard evaluation setup closely follow \citet{gao-etal-2021-making}, which measure finetuning performances across five different randomly sampled training data of size  $\mathbf{K}$ to account for finetuning instability on small datasets \cite{dodge2020fine, mosbach2020stability}. We perform five data subsampling for each dataset and each data size $\mathbf{K}$, where $\mathbf{K}\in \{16,32,64,128,256,512\}$. Note that $\mathbf{K}$ indicates the number of examples \textit{per label}. We use the original development sets of each training dataset for testing the \textit{in-distribution} performance. We perform all experiments using the \texttt{RoBERTa-large} model \cite{liu2019roberta}. 


\paragraph{Inference heuristics across data regimes} We show the results of the prompt-based finetuning across different $\mathbf{K}$ in Figure \ref{fig:data_regimes}. For the in-distribution evaluations (leftmost of each plot), the prompt-based models finetuned on MNLI, SNLI, and QQP improve rapidly with more training data before saturating at $\mathbf{K}=512$. In contrast to the in-distribution results, we observe a different trajectory of performance on the three challenge datasets. On the Scramble and HANS sets, prompt-based models show non-trivial zero-shot performance ($\mathbf{K}=0$) that is above its in-distribution counterpart. However, as more data is available, the models exhibit stronger indication of adopting heuristics. Namely, the performance on examples subset that \textit{support} the heuristics increases, while the performance on cases that are \textit{against} heuristics decreases. This pattern is most pronounced on the lexical overlap subset of HANS, where the median accuracy on non-entailment subset drops to below 10\% while the entailment performance reaches 100\%.
The results suggest that 
few-shot finetuning can be destructive against the initial ability of prompt-based classifier to ignore surface features like lexical overlap. Finetuning appears to over-adjust model parameters to the small target data, which contain very few to no 
counter-examples to the heuristics
\cite{min-etal-2020-syntactic, Lovering2020Predicting}. 





\section{Avoiding Inference Heuristics}
Here we look to mitigate the adverse impact of finetuning by viewing the issue as an instance of catastrophic forgetting \cite{ French1999CatastrophicFI}, which is characterized by the loss of performance on the original dataset after subsequent finetuning on new data. We then propose a regularized prompt-based finetuning based on the Elastic Weight Consolidation (EWC) method \cite{Kirkpatrick2017OvercomingCF}, which penalizes updates on weights crucial for the original zero-shot performance. EWC identifies these weights using empirical Fisher matrix \cite{Martens2020New}, which requires samples of the original dataset. To omit the need of accessing the pretraining data, we follow \citet{chen-etal-2020-recall} that assume stronger independence between the Fisher information and the corresponding weights. The penalty term is now akin to the L2 loss between updated weights $\theta_i$ and the original weights $\theta_i^{*}$, resulting in the following overall loss:
$$
\mathbf{L}_{rFT} = \alpha \mathbf{L}_{FT} + (1-\alpha) \frac{\lambda}{2}\sum_i (\theta_i - \theta_i^{*})^2
$$
where $\mathbf{L}_{FT}$ is a standard cross entropy, $\lambda$ is a quadratic penalty coefficient, and $\alpha$ is a coefficient to linearly combine the two terms. We use the RecAdam implementation \cite{chen-etal-2020-recall} for this loss, which also applies an annealing mechanism to gradually upweight the standard loss $\mathbf{L}_{FT}$ toward the end of training.\footnote{See Appendix \ref{implementation} for implementation details.}

\begin{table*}
\centering
\footnotesize
\begin{tabular}{r|ccc|ccc|ccc}
\toprule
                 & \multicolumn{3}{c|}{\textbf{MNLI} (acc.)} & \multicolumn{3}{c|}{\textbf{QQP} (F1)}   & \multicolumn{3}{c}{\textbf{SNLI} (acc.)}     \\
                 & \textbf{In-dist.} & \textit{HANS} & \textbf{avg.}             & \textbf{In-dist.} & \textit{PAWS} & \textbf{avg.}& \textbf{In-dist.} & \textit{Scramble}& \textbf{avg.}  \\
\midrule
\multicolumn{10}{c}{\textbf{Prompt-based}}\\
\midrule
\textit{zero-shot} \#0       &  51.1    &            62.6       & 56.8  & 35.4 &     51.8  & 43.6 &  49.7    &    64.7   & 57.2   \\
\midrule
FT \#512       &   84.3   &         54.8          & 69.5 &   82.1   &  29.6  & 55.8 &   88.1   &     50.1  &   69.1   \\
\midrule
\textbf{rFT \#512}       &  82.7    &           60.2 &    \textbf{71.5}     &   81.5  & 37.1  & \textbf{59.3}  &    87.6  &    55.4    & \textbf{71.5} \\
\midrule
FT-fix18 \#512  &   76.5   &        61.6        &  69.1 &    78.6 &  35.6 &   57.1 &   84.5   & 45.3  & 64.9 \\
FT-fix12 \#512  &  83.5    &        54.3    &        68.9 &  81.9    & 35.3  &  57.1  &   87.1   &    50.5  & 68.8   \\
FT-fix6  \#512  &   84.2   &        52.9       &    68.5  &  82.1   & 32.7    & 57.4 &   87.9  &       50.1     & 68.9  \\
\midrule
\multicolumn{10}{c}{\textbf{Classifier head}}\\
\midrule
FT \#512 &   81.4   &      52.6     &   67.0   &   80.9    & 26.8   & 53.8  &   86.5  &    49.8  & 68.1 \\
\bottomrule          
\end{tabular}
\caption{Results of different strategies for finetuning prompt-based model (using \#$k$ examples). Models are evaluated against the in-distribution set and corresponding challenge sets. The zero-shot row indicates prompting results before finetuning. The \textit{avg} columns report the average score on in-distribution and challenge datasets.}
\label{tab:results}
\end{table*}

\paragraph{Baselines} We compare regularized finetuning with another method that also minimally update the pretraining weights. We consider simple weight fixing of the first $n$ layers of the pretrained model, where the $n$ layers are frozen (including the token embeddings) and only the weights of upper layers and LM head are updated throughout the finetuning. In the evaluation, we use $n\in \{6,12,18\}$. We refer to these baselines as FT-fix$n$. 


\paragraph{Results} We evaluate all the considered finetuning strategies by taking their median performance after finetuning on 512 examples (for each label) and compare them with the original zero-shot performance. We report the results on Table \ref{tab:results}, which also include the results of standard classifier head finetuning (last row). We observe the following: (1) Freezing the layers has mixed challenge set results, e.g., FT-fix18 improves over vanilla prompt-based finetuning on HANS and PAWS, but degrades Scramble and all in-distribution performances; (2) The L2 regularization strategy, rFT, achieves consistent improvements on the challenge sets while only costs small drop on the corresponding in-distribution performance, e.g., +6pp, +8pp, and +5pp on HANS, PAWS, and Scramble, respectively; (3) Although vanilla \textit{prompt-based} finetuning performs relatively poorly, it still has an advantage over standard \textit{classifier head} finetuning by +2.5pp, +2.0pp, and +1.0pp on the average scores of each in-distribution and challenge dataset pair.


\begin{figure}%
\centering
\includegraphics[height=5.2cm]{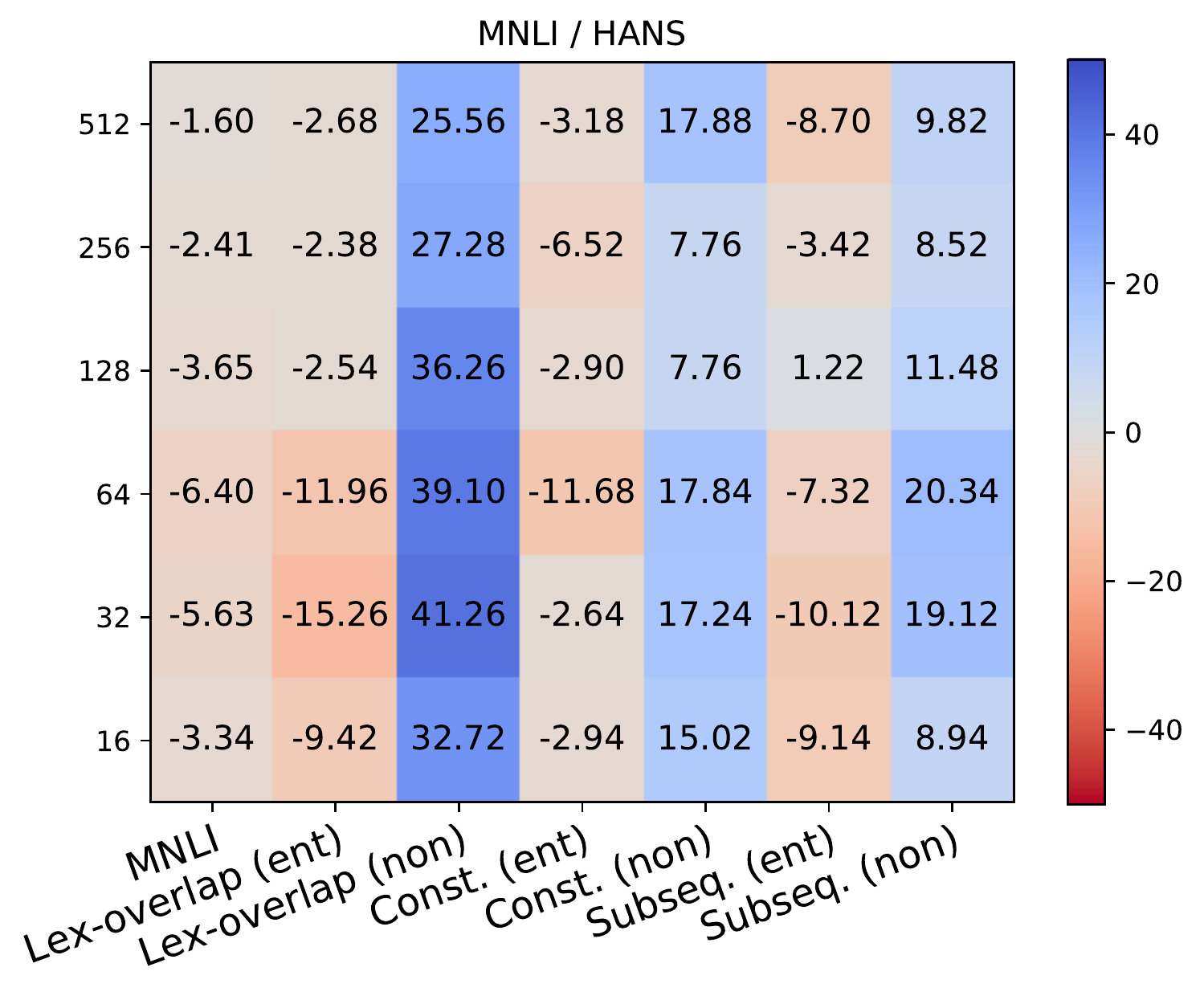}
\caption{Relative difference between median accuracy of prompt-based finetuning across data regimes (y axis) with and without regularization on MNLI and HANS.}
\label{fig:ewc_results1}
\end{figure}

Additionally, Figure \ref{fig:ewc_results1} shows rFT's improvement over vanilla prompt-based finetuning across data regimes on MNLI and HANS. We observe that the advantage of rFT is the strongest on the lexical overlap subset, which initially shows the highest zero-shot performance. The results also suggest that the benefit of rFT peaks at mid data regimes (e.g., $\mathbf{K}=32$), before saturating when training data size is increased further. We also note that our results are consistent when we evaluate alternative prompt templates, or finetune for varying number of epochs.\footnote{See Appendix \ref{add_results} for the detailed results.} The latter indicates that the adoption of inference heuristics is more likely attributed to the amount of training examples rather than the number of learning steps.






\section{Related Work}
\paragraph{Inference Heuristics} Our work relates to a large body of literature on the problem of ``bias'' in the training datasets and the ramifications to the resulting models across various language understanding tasks \cite{niven-kao-2019-probing, poliak-etal-2020-hypothesis, tsuchiya-2018-performance, gururangan-etal-2020-dont}. Previous work shows that the artifacts of data annotations result in spurious surface cues, which gives away the labels, allowing models to perform well without properly learning the intended task. For instance, models are shown to adopt heuristics based on the presence of certain indicative words or phrases in tasks such as reading comprehension \cite{kaushik-lipton-2018-much}, story cloze completion \cite{schwartz-etal-2017-effect, cai-etal-2017-pay}, fact verification \cite{schuster-etal-2019-towards}, argumentation mining \cite{niven-kao-2019-probing}, and natural language inference \cite{gururangan-etal-2020-dont}. Heuristics in models are often investigated using constructed ``challenge datasets'' consisting of counter-examples to the spurious cues, which mostly result in incorrect predictions \cite{jia-liang-2017-adversarial, glockner-etal-2018-breaking, naik-etal-2018-stress, mccoy-etal-2019-right}. Although the problem has been extensively studied, most works focus on models that are trained in standard settings where larger training datasets are available. Our work provides new insights in inference heuristics in models that are trained in zero- and few-shot settings. 

\paragraph{Heuristics Mitigation} Significant prior work attempt to mitigate the heuristics in models by improving the training dataset. \citet{zellers-etal-2019-hellaswag, sakaguchi-etal-2020-winogrande} propose to reduce artifacts in the training data by using adversarial filtering methods; \citet{nie-etal-2020-adversarial,Kaushik2020Learning} aim at a similar improvement via iterative data collection using human-in-the-loop; \citet{min-etal-2020-syntactic, schuster-etal-2021-get, liu-etal-2019-inoculation, rozen-etal-2019-diversify} augment the training dataset with adversarial instances; and \citet{moosavi-etal-2020-improving} augment each training instances with their semantic roles information. Complementary to this, recent work introduces various learning algorithms to avoid adopting heuristics including by re-weighting \cite{he-etal-2019-unlearn, karimi-mahabadi-etal-2020-end, clark-etal-2020-learning} or regularizing the confidence \cite{utama-etal-2020-mind, du-etal-2021-interpreting} on the training instances which exhibit certain biases.
The type of bias can be identified automatically \cite{yaghoobzadeh-etal-2021-increasing, utama-etal-2020-towards, sanh-etal-2021-learning, clark-etal-2020-learning} or by hand-crafted models designed based on prior knowledge about the bias. Our finding suggests that prompted zero-shot models are less reliant on heuristics when tested against examples containing the cues,
and preserving this learned behavior is crucial to obtain more robust finetuned models.

\paragraph{Efficiency and Robustness} Prompting formulation enables language models to learn efficiently from a small number of training examples, which in turn reduces the computational cost for training \cite{le-scao-rush-2021-many}. The efficiency benefit from prompting is very relevant to the larger efforts towards sustainable and green NLP models \cite{sustainlp-2020-sustainlp, schwartz-etal-2020-green-ai} which encompass a flurry of techniques including knowledge distillation \cite{hinton-etal-2015-distilling, sanh-etal-2019-distilbert}, pruning \cite{han-etal-2015-learning}, quantization \cite{Jacob_2018_CVPR}, and early exiting \cite{schwartz-etal-2020-right, xin-etal-2020-early}. Recently, \citet{hooker-etal-2020-characterising} show that methods improving compute and memory efficiency using pruning and quantization may be at odds with robustness and fairness. They report that while performance on standard test sets is largely unchanged, the performance of efficient models on certain underrepresented subsets of the data is disproportionately reduced, suggesting the importance of a more comprehensive evaluation to estimate overall changes in performance. 





\section{Conclusion}
Our experiments shed light on
the negative impact of low resource finetuning to the models' overall performance that is previously obscured by standard evaluation setup. The results indicate that while finetuning helps prompt-based models to rapidly gain the \textit{in-distribution} improvement as more labeled data are available, it also gradually increases models' reliance on \textit{surface heuristics}, which we show to be less present in the zero-shot evaluation. We further demonstrate that applying regularization that preserves pretrained weights during finetuning mitigates the adoption of heuristics while also maintains high in-distribution performances.

\section*{Acknowledgement}
We thank Michael Bugert, Tim Baumgärtner, Jan Buchman, and the anonymous reviewers for their constructive feedback. This work is funded by the German Research Foundation through the research training group AIPHES (GRK 1994/1) and  by the German Federal Ministry of Education and Research and
the Hessen State Ministry for Higher Education, Research and the Arts within their joint support of the National Research Center for Applied Cybersecurity ATHENE.

\bibliography{anthology,custom,ref}

\begin{thebibliography}{61}
\expandafter\ifx\csname natexlab\endcsname\relax\def\natexlab#1{#1}\fi

\bibitem[{Bowman et~al.(2015)Bowman, Angeli, Potts, and
  Manning}]{bowman-etal-2015-large}
Samuel~R. Bowman, Gabor Angeli, Christopher Potts, and Christopher~D. Manning.
  2015.
\newblock \href {https://doi.org/10.18653/v1/D15-1075} {A large annotated
  corpus for learning natural language inference}.
\newblock In \emph{Proceedings of the 2015 Conference on Empirical Methods in
  Natural Language Processing}, pages 632--642, Lisbon, Portugal. Association
  for Computational Linguistics.

\bibitem[{Cai et~al.(2017)Cai, Tu, and Gimpel}]{cai-etal-2017-pay}
Zheng Cai, Lifu Tu, and Kevin Gimpel. 2017.
\newblock \href {https://doi.org/10.18653/v1/P17-2097} {Pay attention to the
  ending:strong neural baselines for the {ROC} story cloze task}.
\newblock In \emph{Proceedings of the 55th Annual Meeting of the Association
  for Computational Linguistics (Volume 2: Short Papers)}, pages 616--622,
  Vancouver, Canada. Association for Computational Linguistics.

\bibitem[{Chen et~al.(2020)Chen, Hou, Cui, Che, Liu, and
  Yu}]{chen-etal-2020-recall}
Sanyuan Chen, Yutai Hou, Yiming Cui, Wanxiang Che, Ting Liu, and Xiangzhan Yu.
  2020.
\newblock \href {https://doi.org/10.18653/v1/2020.emnlp-main.634} {Recall and
  learn: Fine-tuning deep pretrained language models with less forgetting}.
\newblock In \emph{Proceedings of the 2020 Conference on Empirical Methods in
  Natural Language Processing (EMNLP)}, pages 7870--7881, Online. Association
  for Computational Linguistics.

\bibitem[{Clark et~al.(2020)Clark, Yatskar, and
  Zettlemoyer}]{clark-etal-2020-learning}
Christopher Clark, Mark Yatskar, and Luke Zettlemoyer. 2020.
\newblock \href {https://doi.org/10.18653/v1/2020.findings-emnlp.272} {Learning
  to model and ignore dataset bias with mixed capacity ensembles}.
\newblock In \emph{Findings of the Association for Computational Linguistics:
  EMNLP 2020}, pages 3031--3045, Online. Association for Computational
  Linguistics.

\bibitem[{Dasgupta et~al.(2018)Dasgupta, Guo, Stuhlm{\"{u}}ller, Gershman, and
  Goodman}]{dasgupta2018evaluating}
Ishita Dasgupta, Demi Guo, Andreas Stuhlm{\"{u}}ller, Samuel Gershman, and
  Noah~D. Goodman. 2018.
\newblock \href {https://mindmodeling.org/cogsci2018/papers/0307/index.html}
  {Evaluating compositionality in sentence embeddings}.
\newblock In \emph{Proceedings of the 40th Annual Meeting of the Cognitive
  Science Society, CogSci 2018, Madison, WI, USA, July 25-28, 2018}.
  cognitivesciencesociety.org.

\bibitem[{Dodge et~al.(2020)Dodge, Ilharco, Schwartz, Farhadi, Hajishirzi, and
  Smith}]{dodge2020fine}
Jesse Dodge, Gabriel Ilharco, Roy Schwartz, Ali Farhadi, Hannaneh Hajishirzi,
  and Noah Smith. 2020.
\newblock Fine-tuning pretrained language models: Weight initializations, data
  orders, and early stopping.
\newblock \emph{arXiv preprint arXiv:2002.06305}.

\bibitem[{Du et~al.(2021)Du, Manjunatha, Jain, Deshpande, Dernoncourt, Gu, Sun,
  and Hu}]{du-etal-2021-interpreting}
Mengnan Du, Varun Manjunatha, Rajiv Jain, Ruchi Deshpande, Franck Dernoncourt,
  Jiuxiang Gu, Tong Sun, and Xia Hu. 2021.
\newblock \href {http://arxiv.org/abs/2103.06922} {Towards interpreting and
  mitigating shortcut learning behavior of {NLU} models}.
\newblock \emph{arXiv preprint arXiv:2103.06922}.

\bibitem[{French(1999)}]{French1999CatastrophicFI}
R.~French. 1999.
\newblock Catastrophic forgetting in connectionist networks.
\newblock \emph{Trends in Cognitive Sciences}, 3:128--135.

\bibitem[{Gao et~al.(2021)Gao, Fisch, and Chen}]{gao-etal-2021-making}
Tianyu Gao, Adam Fisch, and Danqi Chen. 2021.
\newblock \href {https://doi.org/10.18653/v1/2021.acl-long.295} {Making
  pre-trained language models better few-shot learners}.
\newblock In \emph{Proceedings of the 59th Annual Meeting of the Association
  for Computational Linguistics and the 11th International Joint Conference on
  Natural Language Processing (Volume 1: Long Papers)}, pages 3816--3830,
  Online. Association for Computational Linguistics.

\bibitem[{Glockner et~al.(2018)Glockner, Shwartz, and
  Goldberg}]{glockner-etal-2018-breaking}
Max Glockner, Vered Shwartz, and Yoav Goldberg. 2018.
\newblock \href {https://doi.org/10.18653/v1/P18-2103} {Breaking {NLI} systems
  with sentences that require simple lexical inferences}.
\newblock In \emph{Proceedings of the 56th Annual Meeting of the Association
  for Computational Linguistics (Volume 2: Short Papers)}, pages 650--655,
  Melbourne, Australia. Association for Computational Linguistics.

\bibitem[{Gururangan et~al.(2020)Gururangan, Marasovi{\'c}, Swayamdipta, Lo,
  Beltagy, Downey, and Smith}]{gururangan-etal-2020-dont}
Suchin Gururangan, Ana Marasovi{\'c}, Swabha Swayamdipta, Kyle Lo, Iz~Beltagy,
  Doug Downey, and Noah~A. Smith. 2020.
\newblock \href {https://doi.org/10.18653/v1/2020.acl-main.740} {Don{'}t stop
  pretraining: Adapt language models to domains and tasks}.
\newblock In \emph{Proceedings of the 58th Annual Meeting of the Association
  for Computational Linguistics}, pages 8342--8360, Online. Association for
  Computational Linguistics.

\bibitem[{Han et~al.(2015)Han, Pool, Tran, and Dally}]{han-etal-2015-learning}
Song Han, Jeff Pool, John Tran, and William~J. Dally. 2015.
\newblock Learning both weights and connections for efficient neural networks.
\newblock In \emph{Proceedings of the 28th International Conference on Neural
  Information Processing Systems - Volume 1}, NIPS'15, page 1135–1143,
  Cambridge, MA, USA. MIT Press.

\bibitem[{He et~al.(2019)He, Zha, and Wang}]{he-etal-2019-unlearn}
He~He, Sheng Zha, and Haohan Wang. 2019.
\newblock \href {https://doi.org/10.18653/v1/D19-6115} {Unlearn dataset bias in
  natural language inference by fitting the residual}.
\newblock In \emph{Proceedings of the 2nd Workshop on Deep Learning Approaches
  for Low-Resource NLP (DeepLo 2019)}, pages 132--142, Hong Kong, China.
  Association for Computational Linguistics.

\bibitem[{Hewitt and Manning(2019)}]{hewitt-manning-2019-structural}
John Hewitt and Christopher~D. Manning. 2019.
\newblock \href {https://doi.org/10.18653/v1/N19-1419} {{A} structural probe
  for finding syntax in word representations}.
\newblock In \emph{Proceedings of the 2019 Conference of the North {A}merican
  Chapter of the Association for Computational Linguistics: Human Language
  Technologies, Volume 1 (Long and Short Papers)}, pages 4129--4138,
  Minneapolis, Minnesota. Association for Computational Linguistics.

\bibitem[{Hinton et~al.(2015)Hinton, Vinyals, and
  Dean}]{hinton-etal-2015-distilling}
Geoffrey Hinton, Oriol Vinyals, and Jeffrey Dean. 2015.
\newblock \href {http://arxiv.org/abs/1503.02531} {Distilling the knowledge in
  a neural network}.
\newblock In \emph{NeurIPS Deep Learning and Representation Learning Workshop}.

\bibitem[{Hooker et~al.(2020)Hooker, Moorosi, Clark, Bengio, and
  Denton}]{hooker-etal-2020-characterising}
Sara Hooker, Nyalleng Moorosi, Gregory Clark, Samy Bengio, and Emily Denton.
  2020.
\newblock \href {http://arxiv.org/abs/2010.03058} {Characterising bias in
  compressed models}.
\newblock \emph{arXiv preprint arXiv:2010.03058}.

\bibitem[{Jacob et~al.(2018)Jacob, Kligys, Chen, Zhu, Tang, Howard, Adam, and
  Kalenichenko}]{Jacob_2018_CVPR}
Benoit Jacob, Skirmantas Kligys, Bo~Chen, Menglong Zhu, Matthew Tang, Andrew
  Howard, Hartwig Adam, and Dmitry Kalenichenko. 2018.
\newblock Quantization and training of neural networks for efficient
  integer-arithmetic-only inference.
\newblock In \emph{Proceedings of the IEEE Conference on Computer Vision and
  Pattern Recognition (CVPR)}.

\bibitem[{Jia and Liang(2017)}]{jia-liang-2017-adversarial}
Robin Jia and Percy Liang. 2017.
\newblock \href {https://doi.org/10.18653/v1/D17-1215} {Adversarial examples
  for evaluating reading comprehension systems}.
\newblock In \emph{Proceedings of the 2017 Conference on Empirical Methods in
  Natural Language Processing}, pages 2021--2031, Copenhagen, Denmark.
  Association for Computational Linguistics.

\bibitem[{Karimi~Mahabadi et~al.(2020)Karimi~Mahabadi, Belinkov, and
  Henderson}]{karimi-mahabadi-etal-2020-end}
Rabeeh Karimi~Mahabadi, Yonatan Belinkov, and James Henderson. 2020.
\newblock \href {https://doi.org/10.18653/v1/2020.acl-main.769} {End-to-end
  bias mitigation by modelling biases in corpora}.
\newblock In \emph{Proceedings of the 58th Annual Meeting of the Association
  for Computational Linguistics}, pages 8706--8716, Online. Association for
  Computational Linguistics.

\bibitem[{Kaushik et~al.(2020)Kaushik, Hovy, and Lipton}]{Kaushik2020Learning}
Divyansh Kaushik, Eduard Hovy, and Zachary Lipton. 2020.
\newblock \href {https://openreview.net/forum?id=Sklgs0NFvr} {Learning the
  difference that makes a difference with counterfactually-augmented data}.
\newblock In \emph{8th International Conference on Learning Representations,
  {ICLR} 2020, Virtual Conference, 26 April - 1 May, 2019}. OpenReview.net.

\bibitem[{Kaushik and Lipton(2018)}]{kaushik-lipton-2018-much}
Divyansh Kaushik and Zachary~C. Lipton. 2018.
\newblock \href {https://doi.org/10.18653/v1/D18-1546} {How much reading does
  reading comprehension require? a critical investigation of popular
  benchmarks}.
\newblock In \emph{Proceedings of the 2018 Conference on Empirical Methods in
  Natural Language Processing}, pages 5010--5015, Brussels, Belgium.
  Association for Computational Linguistics.

\bibitem[{Kirkpatrick et~al.(2017)Kirkpatrick, Pascanu, Rabinowitz, Veness,
  Desjardins, Rusu, Milan, Quan, Ramalho, Grabska-Barwinska, Hassabis, Clopath,
  Kumaran, and Hadsell}]{Kirkpatrick2017OvercomingCF}
J.~Kirkpatrick, Razvan Pascanu, Neil~C. Rabinowitz, J.~Veness, G.~Desjardins,
  Andrei~A. Rusu, K.~Milan, John Quan, Tiago Ramalho, Agnieszka
  Grabska-Barwinska, D.~Hassabis, C.~Clopath, D.~Kumaran, and R.~Hadsell. 2017.
\newblock Overcoming catastrophic forgetting in neural networks.
\newblock \emph{Proceedings of the National Academy of Sciences}, 114:3521 --
  3526.

\bibitem[{Le~Scao and Rush(2021)}]{le-scao-rush-2021-many}
Teven Le~Scao and Alexander Rush. 2021.
\newblock \href {https://doi.org/10.18653/v1/2021.naacl-main.208} {How many
  data points is a prompt worth?}
\newblock In \emph{Proceedings of the 2021 Conference of the North American
  Chapter of the Association for Computational Linguistics: Human Language
  Technologies}, pages 2627--2636, Online. Association for Computational
  Linguistics.

\bibitem[{Liu et~al.(2019{\natexlab{a}})Liu, Schwartz, and
  Smith}]{liu-etal-2019-inoculation}
Nelson~F. Liu, Roy Schwartz, and Noah~A. Smith. 2019{\natexlab{a}}.
\newblock \href {https://doi.org/10.18653/v1/N19-1225} {Inoculation by
  fine-tuning: A method for analyzing challenge datasets}.
\newblock In \emph{Proceedings of the 2019 Conference of the North {A}merican
  Chapter of the Association for Computational Linguistics: Human Language
  Technologies, Volume 1 (Long and Short Papers)}, pages 2171--2179,
  Minneapolis, Minnesota. Association for Computational Linguistics.

\bibitem[{Liu et~al.(2019{\natexlab{b}})Liu, Ott, Goyal, Du, Joshi, Chen, Levy,
  Lewis, Zettlemoyer, and Stoyanov}]{liu2019roberta}
Yinhan Liu, Myle Ott, Naman Goyal, Jingfei Du, Mandar Joshi, Danqi Chen, Omer
  Levy, Mike Lewis, Luke Zettlemoyer, and Veselin Stoyanov. 2019{\natexlab{b}}.
\newblock \href {http://arxiv.org/abs/1907.11692} {{RoBERTa}: {A} robustly
  optimized {BERT} pretraining approach}.
\newblock \emph{arXiv preprint arXiv:1907.11692}.

\bibitem[{Lovering et~al.(2021)Lovering, Jha, Linzen, and
  Pavlick}]{Lovering2020Predicting}
Charles Lovering, Rohan Jha, Tal Linzen, and Ellie Pavlick. 2021.
\newblock \href {https://openreview.net/pdf?id=mNtmhaDkAr} {Predicting
  inductive biases of pre-trained models}.
\newblock In \emph{International Conference on Learning Representations, {ICLR}
  2021, Virtual Conference, 3 May - 8 May, 2021}. OpenReview.net.

\bibitem[{Martens(2020)}]{Martens2020New}
James Martens. 2020.
\newblock \href {http://jmlr.org/papers/v21/17-678.html} {New insights and
  perspectives on the natural gradient method}.
\newblock \emph{Journal of Machine Learning Research}, 21(146):1--76.

\bibitem[{McCoy et~al.(2019)McCoy, Pavlick, and Linzen}]{mccoy-etal-2019-right}
Tom McCoy, Ellie Pavlick, and Tal Linzen. 2019.
\newblock \href {https://doi.org/10.18653/v1/P19-1334} {Right for the wrong
  reasons: Diagnosing syntactic heuristics in natural language inference}.
\newblock In \emph{Proceedings of the 57th Annual Meeting of the Association
  for Computational Linguistics}, pages 3428--3448, Florence, Italy.
  Association for Computational Linguistics.

\bibitem[{Min et~al.(2020)Min, McCoy, Das, Pitler, and
  Linzen}]{min-etal-2020-syntactic}
Junghyun Min, R.~Thomas McCoy, Dipanjan Das, Emily Pitler, and Tal Linzen.
  2020.
\newblock \href {https://doi.org/10.18653/v1/2020.acl-main.212} {Syntactic data
  augmentation increases robustness to inference heuristics}.
\newblock In \emph{Proceedings of the 58th Annual Meeting of the Association
  for Computational Linguistics}, pages 2339--2352, Online. Association for
  Computational Linguistics.

\bibitem[{Moosavi et~al.(2020{\natexlab{a}})Moosavi, de~Boer, Utama, and
  Gurevych}]{moosavi-etal-2020-improving}
Nafise~Sadat Moosavi, Marcel de~Boer, Prasetya~Ajie Utama, and Iryna Gurevych.
  2020{\natexlab{a}}.
\newblock \href {http://arxiv.org/abs/2010.12510} {Improving robustness by
  augmenting training sentences with predicate-argument structures}.
\newblock \emph{arXiv preprint arXiv:2010.12510}.

\bibitem[{Moosavi et~al.(2020{\natexlab{b}})Moosavi, Fan, Shwartz,
  Glava{\v{s}}, Joty, Wang, and Wolf}]{sustainlp-2020-sustainlp}
Nafise~Sadat Moosavi, Angela Fan, Vered Shwartz, Goran Glava{\v{s}}, Shafiq
  Joty, Alex Wang, and Thomas Wolf, editors. 2020{\natexlab{b}}.
\newblock \href {https://www.aclweb.org/anthology/2020.sustainlp-1.0}
  {\emph{Proceedings of SustaiNLP: Workshop on Simple and Efficient Natural
  Language Processing}}. Association for Computational Linguistics, Online.

\bibitem[{Mosbach et~al.(2021)Mosbach, Andriushchenko, and
  Klakow}]{mosbach2020stability}
Marius Mosbach, Maksym Andriushchenko, and Dietrich Klakow. 2021.
\newblock \href {https://openreview.net/forum?id=nzpLWnVAyah} {On the stability
  of fine-tuning {BERT:} misconceptions, explanations, and strong baselines}.
\newblock In \emph{9th International Conference on Learning Representations,
  {ICLR} 2021, Virtual Event, Austria, May 3-7, 2021}. OpenReview.net.

\bibitem[{Naik et~al.(2018)Naik, Ravichander, Sadeh, Rose, and
  Neubig}]{naik-etal-2018-stress}
Aakanksha Naik, Abhilasha Ravichander, Norman Sadeh, Carolyn Rose, and Graham
  Neubig. 2018.
\newblock \href {https://www.aclweb.org/anthology/C18-1198} {Stress test
  evaluation for natural language inference}.
\newblock In \emph{Proceedings of the 27th International Conference on
  Computational Linguistics}, pages 2340--2353, Santa Fe, New Mexico, USA.
  Association for Computational Linguistics.

\bibitem[{Nie et~al.(2019)Nie, Wang, and Bansal}]{Nie_Wang_Bansal_2019}
Yixin Nie, Yicheng Wang, and Mohit Bansal. 2019.
\newblock \href {https://doi.org/10.1609/aaai.v33i01.33016867} {Analyzing
  compositionality-sensitivity of nli models}.
\newblock \emph{Proceedings of the AAAI Conference on Artificial Intelligence},
  33(01):6867--6874.

\bibitem[{Nie et~al.(2020)Nie, Williams, Dinan, Bansal, Weston, and
  Kiela}]{nie-etal-2020-adversarial}
Yixin Nie, Adina Williams, Emily Dinan, Mohit Bansal, Jason Weston, and Douwe
  Kiela. 2020.
\newblock \href {https://doi.org/10.18653/v1/2020.acl-main.441} {Adversarial
  {NLI}: A new benchmark for natural language understanding}.
\newblock In \emph{Proceedings of the 58th Annual Meeting of the Association
  for Computational Linguistics}, pages 4885--4901, Online. Association for
  Computational Linguistics.

\bibitem[{Niven and Kao(2019)}]{niven-kao-2019-probing}
Timothy Niven and Hung-Yu Kao. 2019.
\newblock \href {https://doi.org/10.18653/v1/P19-1459} {Probing neural network
  comprehension of natural language arguments}.
\newblock In \emph{Proceedings of the 57th Annual Meeting of the Association
  for Computational Linguistics}, pages 4658--4664, Florence, Italy.
  Association for Computational Linguistics.

\bibitem[{Poliak et~al.(2018)Poliak, Naradowsky, Haldar, Rudinger, and
  Van~Durme}]{poliak-etal-2020-hypothesis}
Adam Poliak, Jason Naradowsky, Aparajita Haldar, Rachel Rudinger, and Benjamin
  Van~Durme. 2018.
\newblock \href {https://doi.org/10.18653/v1/S18-2023} {Hypothesis only
  baselines in natural language inference}.
\newblock In \emph{Proceedings of the Seventh Joint Conference on Lexical and
  Computational Semantics}, pages 180--191, New Orleans, Louisiana. Association
  for Computational Linguistics.

\bibitem[{Radford et~al.(2019)Radford, Wu, Child, Luan, Amodei, and
  Sutskever}]{radford2019language}
Alec Radford, Jeff Wu, Rewon Child, David Luan, Dario Amodei, and Ilya
  Sutskever. 2019.
\newblock Language models are unsupervised multitask learners.
\newblock Technical report, OpenAI.

\bibitem[{Rozen et~al.(2019)Rozen, Shwartz, Aharoni, and
  Dagan}]{rozen-etal-2019-diversify}
Ohad Rozen, Vered Shwartz, Roee Aharoni, and Ido Dagan. 2019.
\newblock \href {https://doi.org/10.18653/v1/K19-1019} {Diversify your
  datasets: Analyzing generalization via controlled variance in adversarial
  datasets}.
\newblock In \emph{Proceedings of the 23rd Conference on Computational Natural
  Language Learning (CoNLL)}, pages 196--205, Hong Kong, China. Association for
  Computational Linguistics.

\bibitem[{Sakaguchi et~al.(2020)Sakaguchi, Bras, Bhagavatula, and
  Choi}]{sakaguchi-etal-2020-winogrande}
Keisuke Sakaguchi, Ronan~Le Bras, Chandra Bhagavatula, and Yejin Choi. 2020.
\newblock \href {https://aaai.org/ojs/index.php/AAAI/article/view/6399}
  {Winogrande: An adversarial winograd schema challenge at scale}.
\newblock In \emph{The Thirty-Fourth {AAAI} Conference on Artificial
  Intelligence, {AAAI} 2020}, pages 8732--8740. {AAAI} Press.

\bibitem[{Sanh et~al.(2019)Sanh, Debut, Chaumond, and
  Wolf}]{sanh-etal-2019-distilbert}
Victor Sanh, Lysandre Debut, Julien Chaumond, and Thomas Wolf. 2019.
\newblock \href {http://arxiv.org/abs/1910.01108} {{DistilBERT}, a distilled
  version of {BERT:} smaller, faster, cheaper and lighter}.
\newblock \emph{arXiv preprint arXiv:1910.01108}.

\bibitem[{Sanh et~al.(2021)Sanh, Wolf, Belinkov, and
  Rush}]{sanh-etal-2021-learning}
Victor Sanh, Thomas Wolf, Yonatan Belinkov, and Alexander~M. Rush. 2021.
\newblock \href {https://openreview.net/forum?id=Hf3qXoiNkR} {Learning from
  others' mistakes: Avoiding dataset biases without modeling them}.
\newblock In \emph{9th International Conference on Learning Representations,
  {ICLR} 2021, Virtual Event, Austria, May 3-7, 2021}. OpenReview.net.

\bibitem[{Schick et~al.(2020)Schick, Schmid, and
  Sch{\"u}tze}]{schick-etal-2020-automatically}
Timo Schick, Helmut Schmid, and Hinrich Sch{\"u}tze. 2020.
\newblock \href {https://doi.org/10.18653/v1/2020.coling-main.488}
  {Automatically identifying words that can serve as labels for few-shot text
  classification}.
\newblock In \emph{Proceedings of the 28th International Conference on
  Computational Linguistics}, pages 5569--5578, Barcelona, Spain (Online).
  International Committee on Computational Linguistics.

\bibitem[{Schick and
  Sch{\"u}tze(2021{\natexlab{a}})}]{schick-schutze-2021-exploiting}
Timo Schick and Hinrich Sch{\"u}tze. 2021{\natexlab{a}}.
\newblock \href {https://www.aclweb.org/anthology/2021.eacl-main.20}
  {Exploiting cloze-questions for few-shot text classification and natural
  language inference}.
\newblock In \emph{Proceedings of the 16th Conference of the European Chapter
  of the Association for Computational Linguistics: Main Volume}, pages
  255--269, Online. Association for Computational Linguistics.

\bibitem[{Schick and
  Sch{\"u}tze(2021{\natexlab{b}})}]{schick-schutze-2021-just}
Timo Schick and Hinrich Sch{\"u}tze. 2021{\natexlab{b}}.
\newblock \href {https://doi.org/10.18653/v1/2021.naacl-main.185} {It{'}s not
  just size that matters: Small language models are also few-shot learners}.
\newblock In \emph{Proceedings of the 2021 Conference of the North American
  Chapter of the Association for Computational Linguistics: Human Language
  Technologies}, pages 2339--2352, Online. Association for Computational
  Linguistics.

\bibitem[{Schuster et~al.(2021)Schuster, Fisch, and
  Barzilay}]{schuster-etal-2021-get}
Tal Schuster, Adam Fisch, and Regina Barzilay. 2021.
\newblock \href {https://doi.org/10.18653/v1/2021.naacl-main.52} {Get your
  vitamin {C}! robust fact verification with contrastive evidence}.
\newblock In \emph{Proceedings of the 2021 Conference of the North American
  Chapter of the Association for Computational Linguistics: Human Language
  Technologies}, pages 624--643, Online. Association for Computational
  Linguistics.

\bibitem[{Schuster et~al.(2019)Schuster, Shah, Yeo, Roberto Filizzola~Ortiz,
  Santus, and Barzilay}]{schuster-etal-2019-towards}
Tal Schuster, Darsh Shah, Yun Jie~Serene Yeo, Daniel Roberto Filizzola~Ortiz,
  Enrico Santus, and Regina Barzilay. 2019.
\newblock \href {https://doi.org/10.18653/v1/D19-1341} {Towards debiasing fact
  verification models}.
\newblock In \emph{Proceedings of the 2019 Conference on Empirical Methods in
  Natural Language Processing and the 9th International Joint Conference on
  Natural Language Processing (EMNLP-IJCNLP)}, pages 3419--3425, Hong Kong,
  China. Association for Computational Linguistics.

\bibitem[{Schwartz et~al.(2020{\natexlab{a}})Schwartz, Dodge, Smith, and
  Etzioni}]{schwartz-etal-2020-green-ai}
Roy Schwartz, Jesse Dodge, Noah~A. Smith, and Oren Etzioni. 2020{\natexlab{a}}.
\newblock \href {https://doi.org/10.1145/3381831} {Green {AI}}.
\newblock \emph{Communications of the ACM}, 63(12):54–63.

\bibitem[{Schwartz et~al.(2017)Schwartz, Sap, Konstas, Zilles, Choi, and
  Smith}]{schwartz-etal-2017-effect}
Roy Schwartz, Maarten Sap, Ioannis Konstas, Leila Zilles, Yejin Choi, and
  Noah~A. Smith. 2017.
\newblock \href {https://doi.org/10.18653/v1/K17-1004} {The effect of different
  writing tasks on linguistic style: A case study of the {ROC} story cloze
  task}.
\newblock In \emph{Proceedings of the 21st Conference on Computational Natural
  Language Learning ({C}o{NLL} 2017)}, pages 15--25, Vancouver, Canada.
  Association for Computational Linguistics.

\bibitem[{Schwartz et~al.(2020{\natexlab{b}})Schwartz, Stanovsky, Swayamdipta,
  Dodge, and Smith}]{schwartz-etal-2020-right}
Roy Schwartz, Gabriel Stanovsky, Swabha Swayamdipta, Jesse Dodge, and Noah~A.
  Smith. 2020{\natexlab{b}}.
\newblock \href {https://doi.org/10.18653/v1/2020.acl-main.593} {The right tool
  for the job: Matching model and instance complexities}.
\newblock In \emph{Proceedings of the 58th Annual Meeting of the Association
  for Computational Linguistics}, pages 6640--6651, Online. Association for
  Computational Linguistics.

\bibitem[{Tenney et~al.(2019)Tenney, Das, and Pavlick}]{tenney-etal-2019-bert}
Ian Tenney, Dipanjan Das, and Ellie Pavlick. 2019.
\newblock \href {https://doi.org/10.18653/v1/P19-1452} {{BERT} rediscovers the
  classical {NLP} pipeline}.
\newblock In \emph{Proceedings of the 57th Annual Meeting of the Association
  for Computational Linguistics}, pages 4593--4601, Florence, Italy.
  Association for Computational Linguistics.

\bibitem[{Tsuchiya(2018)}]{tsuchiya-2018-performance}
Masatoshi Tsuchiya. 2018.
\newblock \href {https://www.aclweb.org/anthology/L18-1239} {Performance impact
  caused by hidden bias of training data for recognizing textual entailment}.
\newblock In \emph{Proceedings of the Eleventh International Conference on
  Language Resources and Evaluation ({LREC} 2018)}, Miyazaki, Japan. European
  Language Resources Association (ELRA).

\bibitem[{Utama et~al.(2020{\natexlab{a}})Utama, Moosavi, and
  Gurevych}]{utama-etal-2020-mind}
Prasetya~Ajie Utama, Nafise~Sadat Moosavi, and Iryna Gurevych.
  2020{\natexlab{a}}.
\newblock \href {https://doi.org/10.18653/v1/2020.acl-main.770} {Mind the
  trade-off: Debiasing {NLU} models without degrading the in-distribution
  performance}.
\newblock In \emph{Proceedings of the 58th Annual Meeting of the Association
  for Computational Linguistics}, pages 8717--8729, Online. Association for
  Computational Linguistics.

\bibitem[{Utama et~al.(2020{\natexlab{b}})Utama, Moosavi, and
  Gurevych}]{utama-etal-2020-towards}
Prasetya~Ajie Utama, Nafise~Sadat Moosavi, and Iryna Gurevych.
  2020{\natexlab{b}}.
\newblock \href {https://doi.org/10.18653/v1/2020.emnlp-main.613} {Towards
  debiasing {NLU} models from unknown biases}.
\newblock In \emph{Proceedings of the 2020 Conference on Empirical Methods in
  Natural Language Processing (EMNLP)}, pages 7597--7610, Online. Association
  for Computational Linguistics.

\bibitem[{Wang et~al.(2018)Wang, Singh, Michael, Hill, Levy, and
  Bowman}]{wang2018glue}
Alex Wang, Amanpreet Singh, Julian Michael, Felix Hill, Omer Levy, and Samuel
  Bowman. 2018.
\newblock \href {https://doi.org/10.18653/v1/W18-5446} {{GLUE}: A multi-task
  benchmark and analysis platform for natural language understanding}.
\newblock In \emph{Proceedings of the 2018 {EMNLP} Workshop {B}lackbox{NLP}:
  Analyzing and Interpreting Neural Networks for {NLP}}, pages 353--355,
  Brussels, Belgium. Association for Computational Linguistics.

\bibitem[{Warstadt et~al.(2020)Warstadt, Zhang, Li, Liu, and
  Bowman}]{warstadt-etal-2020-learning}
Alex Warstadt, Yian Zhang, Xiaocheng Li, Haokun Liu, and Samuel~R. Bowman.
  2020.
\newblock \href {https://doi.org/10.18653/v1/2020.emnlp-main.16} {Learning
  which features matter: {R}o{BERT}a acquires a preference for linguistic
  generalizations (eventually)}.
\newblock In \emph{Proceedings of the 2020 Conference on Empirical Methods in
  Natural Language Processing (EMNLP)}, pages 217--235, Online. Association for
  Computational Linguistics.

\bibitem[{Williams et~al.(2018)Williams, Nangia, and
  Bowman}]{williams-etal-2018-broad}
Adina Williams, Nikita Nangia, and Samuel Bowman. 2018.
\newblock \href {https://doi.org/10.18653/v1/N18-1101} {A broad-coverage
  challenge corpus for sentence understanding through inference}.
\newblock In \emph{Proceedings of the 2018 Conference of the North {A}merican
  Chapter of the Association for Computational Linguistics: Human Language
  Technologies, Volume 1 (Long Papers)}, pages 1112--1122, New Orleans,
  Louisiana. Association for Computational Linguistics.

\bibitem[{Xin et~al.(2020)Xin, Nogueira, Yu, and Lin}]{xin-etal-2020-early}
Ji~Xin, Rodrigo Nogueira, Yaoliang Yu, and Jimmy Lin. 2020.
\newblock \href {https://doi.org/10.18653/v1/2020.sustainlp-1.11} {Early
  exiting {BERT} for efficient document ranking}.
\newblock In \emph{Proceedings of SustaiNLP: Workshop on Simple and Efficient
  Natural Language Processing}, pages 83--88, Online. Association for
  Computational Linguistics.

\bibitem[{Yaghoobzadeh et~al.(2021)Yaghoobzadeh, Mehri, Tachet~des Combes,
  Hazen, and Sordoni}]{yaghoobzadeh-etal-2021-increasing}
Yadollah Yaghoobzadeh, Soroush Mehri, Remi Tachet~des Combes, T.~J. Hazen, and
  Alessandro Sordoni. 2021.
\newblock \href {https://aclanthology.org/2021.eacl-main.291} {Increasing
  robustness to spurious correlations using forgettable examples}.
\newblock In \emph{Proceedings of the 16th Conference of the European Chapter
  of the Association for Computational Linguistics: Main Volume}, pages
  3319--3332, Online. Association for Computational Linguistics.

\bibitem[{Zellers et~al.(2019)Zellers, Holtzman, Bisk, Farhadi, and
  Choi}]{zellers-etal-2019-hellaswag}
Rowan Zellers, Ari Holtzman, Yonatan Bisk, Ali Farhadi, and Yejin Choi. 2019.
\newblock \href {https://doi.org/10.18653/v1/P19-1472} {{H}ella{S}wag: Can a
  machine really finish your sentence?}
\newblock In \emph{Proceedings of the 57th Annual Meeting of the Association
  for Computational Linguistics}, pages 4791--4800, Florence, Italy.
  Association for Computational Linguistics.

\bibitem[{Zhang et~al.(2019)Zhang, Baldridge, and He}]{zhang-etal-2019-paws}
Yuan Zhang, Jason Baldridge, and Luheng He. 2019.
\newblock \href {https://doi.org/10.18653/v1/N19-1131} {{PAWS}: Paraphrase
  adversaries from word scrambling}.
\newblock In \emph{Proceedings of the 2019 Conference of the North {A}merican
  Chapter of the Association for Computational Linguistics: Human Language
  Technologies, Volume 1 (Long and Short Papers)}, pages 1298--1308,
  Minneapolis, Minnesota. Association for Computational Linguistics.

\end{thebibliography}
\bibliographystyle{acl_natbib}

\clearpage

\appendix

\section{Experimental Details}
\label{implementation}
\paragraph{Manual templates and mapping} We use the following prompt templates and word-to-label mapping for the three tasks we evaluate on:

\begin{table}[h]
\footnotesize
\centering
 \begin{tabular}{lr} 
 \toprule
 \textbf{Template} & \textbf{Label Words} \\ [0.5ex] 
 \midrule
 \multicolumn{2}{c}{\textbf{MNLI (manual): entailment, neutral, contradiction}}\\
 \midrule
 $s_1 \texttt{?} \texttt{[MASK]} \textrm{, } s_2$ & Yes, Maybe, No \\
 \midrule
 \multicolumn{2}{c}{\textbf{SNLI (manual): entailment, neutral, contradiction}}\\
 \midrule
 $s_1 \texttt{?} \texttt{[MASK]} \textrm{, } s_2$ & Yes, Maybe, No \\
 \midrule
 \multicolumn{2}{c}{\textbf{QQP (manual): duplicate, non-duplicate}}\\
 \midrule
 $s_1 \texttt{[MASK]} \textrm{, } s_2$ & Yes, No \\
 \midrule
 \multicolumn{2}{c}{\textbf{MNLI (auto): entailment, neutral, contradiction}}\\
 \midrule
 $s_1 \texttt{.} \texttt{[MASK]} \textrm{, you are right , } s_2$ & Fine, Plus, Otherwise \\
 $s_1 \texttt{.} \texttt{[MASK]} \textrm{, you're right , } s_2$ & There, Plus, Otherwise \\
 $s_1 \texttt{.} \texttt{[MASK]} \textrm{ ! } s_2$ & Meaning, Plus, Otherwise \\
 \bottomrule
 \end{tabular}
 \caption{Templates and label words used to finetune and evaluate on MNLI, SNLI, and QQP.}
 \label{tab:templates}
\end{table}

The last 3 rows are automatically generated templates and label words that are shown by \citet{gao-etal-2021-making} to improve the few-shot finetuning further. Note that we use the corresponding task's template when evaluating on the challenge datasets.

\paragraph{Challenge datasets} We provide examples from each challenge datasets considered in our evaluation to illustrate sentence pairs that support or are against the heuristics. Table \ref{tab:challenge} shows examples for HANS, PAWS, and Scramble Test. Following \citet{mccoy-etal-2019-right}, we obtain the probability for the \textit{non-entailment} label by summing the probabilities assigned by models trained on MNLI to the \textit{neutral} and \textit{contradiction} labels. We use the \textit{same-type} subset of Scramble Test \cite{dasgupta2018evaluating} which contain examples of both entailment (\textit{support}) and contradiction (\textit{against}) relations.

\begin{table}
\footnotesize
\centering
 \begin{tabular}{lp{0.33\textwidth}} 
 \toprule
 \multicolumn{2}{c}{\textbf{HANS} \cite{mccoy-etal-2019-right}}\\
 \midrule
 premise & The artists avoided the senators that
thanked the tourists. \\
 hypothesis & The artists avoided the senators. \\
 label & entailment \textbf{(support)}\\
 \midrule
 premise & The managers near the scientist resigned. \\
 hypothesis & The scientist resigned. \\
 label & non-entailment \textbf{(against)}\\
 \midrule\multicolumn{2}{c}{\textbf{PAWS} \cite{zhang-etal-2019-paws}}\\
 \midrule
 S1 & What are the driving rules in Georgia versus Mississippi? \\
 S2 & What are the driving rules in Mississippi versus Georgia? \\
 label & duplicate \textbf{(support)}\\
 \midrule
 S1 & Who pays for Hillary's campaigning for Obama? \\
 S2 & Who pays for Obama's campaigning for Hillary? \\
 label & non-duplicate \textbf{(against)}\\
 \midrule\multicolumn{2}{c}{\textbf{Scramble Test} \cite{dasgupta2018evaluating}}\\
 \midrule
 premise & The woman is more cheerful than the man. \\
 hypothesis & The woman is more cheerful than the man. \\
 label & entailment \textbf{(support)}\\
 \midrule
 premise & The woman is more cheerful than the man. \\
 hypothesis &  The man is more cheerful than the woman. \\
 label & contradiction \textbf{(against)}\\
 \bottomrule
 \end{tabular}
 \caption{Sampled examples from each of the challenge datasets we used for evaluation.}
 \label{tab:challenge}
\end{table}

\paragraph{HANS details} HANS dataset is designed based on the insight that the word overlapping between premise and hypothesis in NLI datasets is spuriously correlated with the \textit{entailment} label. HANS consists of examples in which relying to this correlation leads to incorrect label, i.e., hypotheses are \textit{not entailed} by their word-overlapping premises. HANS is split into three test cases: (a) \textbf{{Lexical overlap}} (e.g., ``\textit{{The doctor} was {paid} by {the actor}}'' $\rightarrow$ ``\textit{The doctor paid the actor}''), (b) \textbf{{Subsequence}} (e.g., ``\textit{The doctor near {the actor danced}}'' $\rightarrow$ ``\textit{The actor danced}''), and (c) \textbf{{Constituent}} (e.g., ``\textit{If {the artist slept}, the actor ran}'' $\rightarrow$ ``\textit{The artist slept}''). Each subset contains both entailment and non-entailment examples that always exhibit word overlap. 

\paragraph{Hyperparameters} Following \citet{schick-schutze-2021-just, schick-schutze-2021-exploiting}, we use a fixed set of hyperparameters for all finetuning: learning rate of $1e^{-5}$, batch size of 8, and maximum length size of 256.

\paragraph{Regularization implementation} We use the RecAdam implementation by \citet{chen-etal-2020-recall} with the following hyperparameters. We set the quadratic penalty $\lambda$ to $5000$, and the linear combination factor $\alpha$ is set dynamically throughout the training according to a sigmoid function schedule, where $\alpha$ at step $t$ is defined as:
$$
\alpha = s(t) = \frac{1}{1+\texttt{exp}(-k \cdot (t-t_0))}
$$
where parameter $k$ regulates the rate of the sigmoid, and $t_0$ sets the point where $s(t)$ goes above $0.5$. We set $k$ to $0.01$ and $t_0$ to $0.6$ of the total training steps.

\begin{figure*}%
\centering
\includegraphics[width=.85\linewidth]{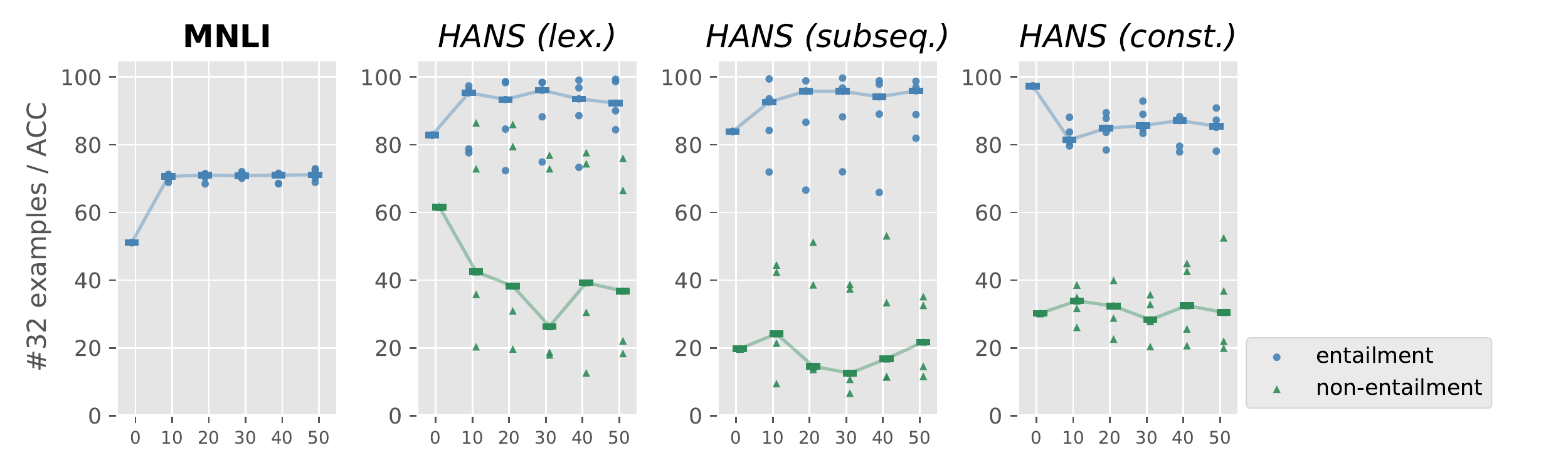}
\includegraphics[width=.85\linewidth]{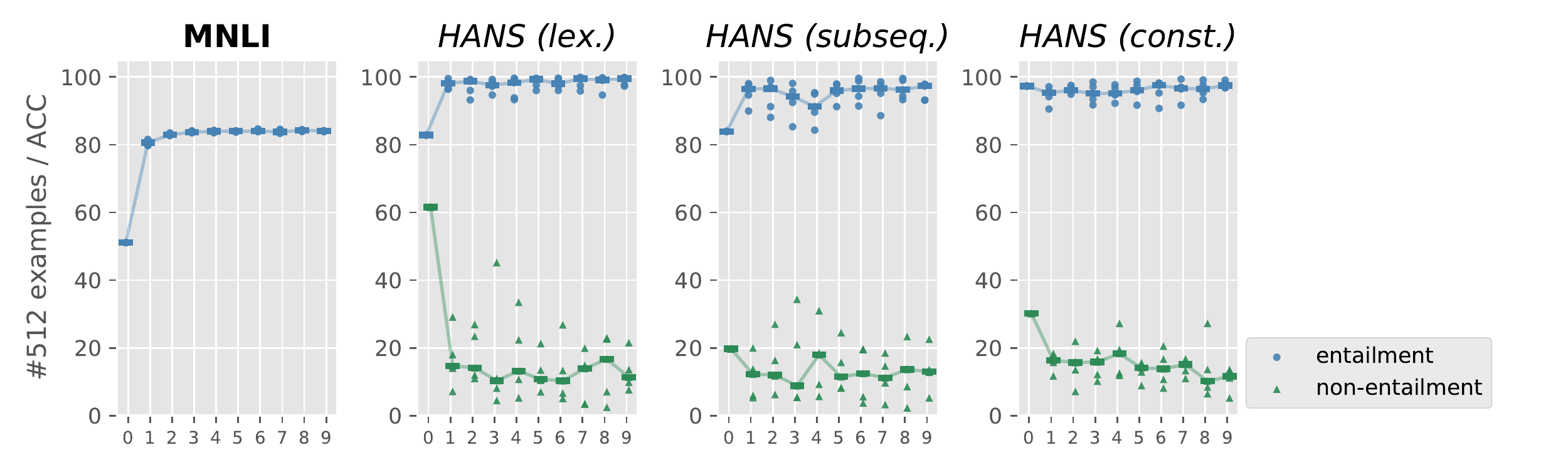}
\caption{Results of prompt-based finetuning with varying number of epochs and fixed amount of training examples. Top: finetuning on 32 examples per label for epochs ranging from 10 to 50. Bottom: finetuning on 512 examples per label for 1 to 9 epochs. Both results show an immediate drop of non-entailment HANS performances which later stagnate even after more training steps.}
\label{fig:epoch_n}
\end{figure*}

\begin{figure}%
\centering
\includegraphics[width=.65\linewidth]{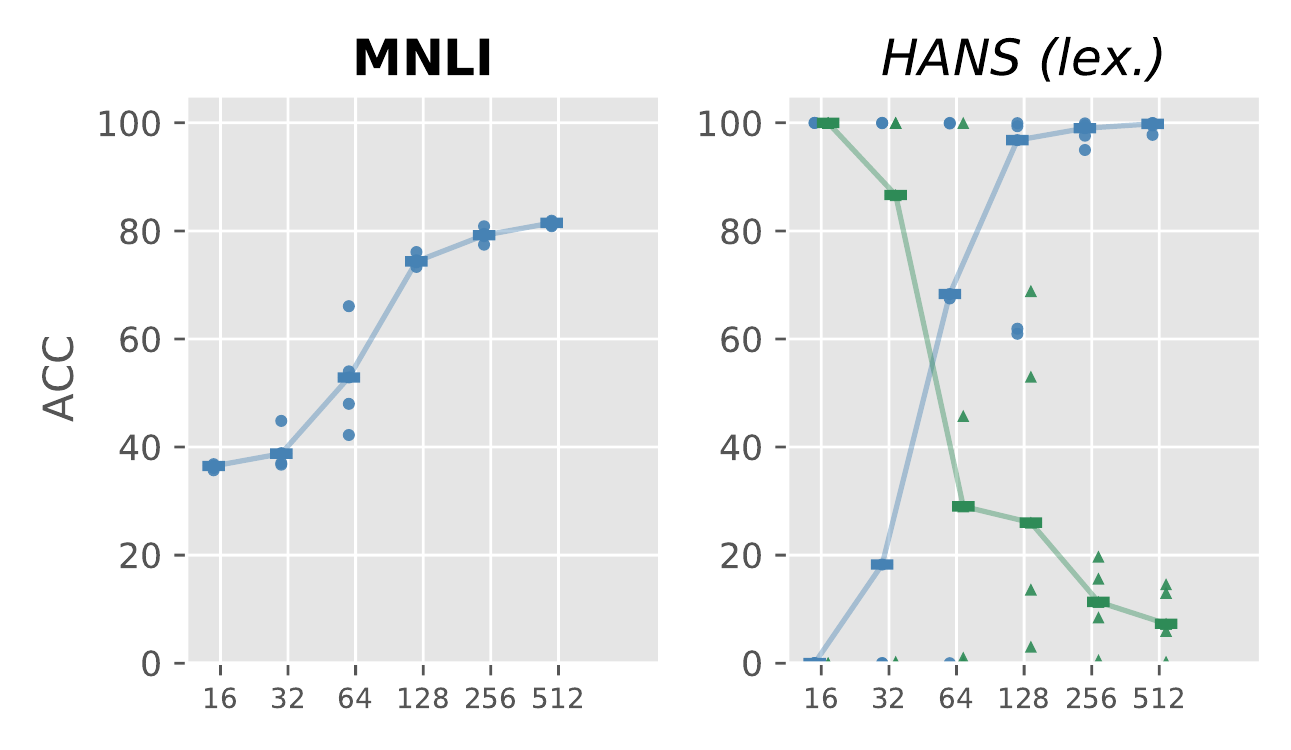}
\includegraphics[width=.65\linewidth]{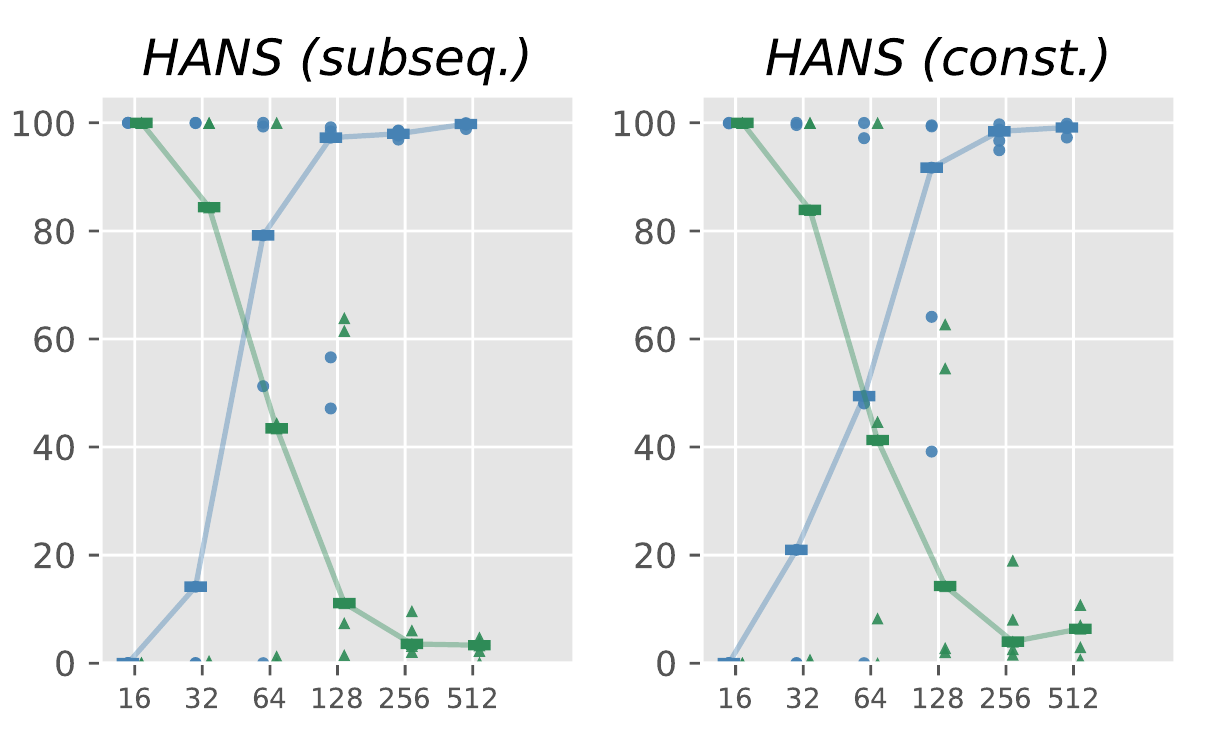}
\caption{Results of non-prompt finetuning.}
\label{fig:fine_tune_data_regimes}
\end{figure}

\section{Additional Results}
\label{add_results}
\paragraph{Standard \texttt{CLS} finetuning} Previously, \citet{gao-etal-2021-making} reported that the performance of standard non-prompt finetuning with additional classifier head (\texttt{CLS}) can converge to that of prompt-based counterpart after certain amount of data, e.g., 512. It is then interesting to compare both finetuning paradigm in terms of their heuristics-related behavior. Figure \ref{fig:fine_tune_data_regimes} shows the results of standard finetuning using standard classifier head across varying data regimes on MNLI and the 3 subsets of HANS. We observe high instability of the results when only small amount of data is available (e.g., $\mathbf{K} = 64$). The learning trajectories are consistent across the HANS subsets, i.e., they start making random predictions on lower data regime and immediately adopt heuristics by predicting almost all examples exhibiting lexical overlap as \textbf{entailment}. We observe that standard prompt-based finetuning still performs better than \texttt{CLS} finetuning, indicating that prompt-based approach provides good initialization to mitigate heuristics, and employing regularization during finetuning can improve the challenge datasets (out-of-distribution) performance further. 
%


\begin{table}
\centering
\footnotesize
\begin{tabular}{l|ll}
\toprule
                 & \multicolumn{2}{c}{\textbf{MNLI} (acc.)}       \\
                 & \textbf{IN} & HANS  \\
\midrule
manual        &  51.1    &            62.6  \\
manual Ft-\#512       &  84.3    &            54.8  \\
\midrule
template-1        &  46.3    &   62.0  \\
template-1 Ft-\#512       &  84.2    &            53.2  \\
\midrule
template-2        &    49.9  &         61.3     \\
template-2 Ft-\#512       &  83.9    &            52.7  \\
\midrule
template-3        &   44.5   &        61.7      \\
template-3 Ft-\#512       &  84.4   &            56.0  \\
\bottomrule          
\end{tabular}
\caption{Evaluation results of different MNLI templates provided by \citet{gao-etal-2021-making}. Models are evaluated against both the in-distribution (\textbf{IN}) set and corresponding challenge set of MNLI.}
\label{tab:templates_results}
\end{table}

\begin{table*}
\centering
\footnotesize
\begin{tabular}{r|ccc|ccc|ccc}
\toprule
                 & \multicolumn{3}{c|}{\textbf{MNLI} (acc.)} & \multicolumn{3}{c|}{\textbf{QQP} (F1)}   & \multicolumn{3}{c}{\textbf{SNLI} (acc.)}     \\
                 & \textbf{In.} & \textit{HANS} & \textbf{avg.}             & \textbf{In.} & \textit{PAWS} & \textbf{avg.}& \textbf{In.} & \textit{Scramble}& \textbf{avg.}  \\
\midrule
zero-shot   \texttt{RoBERTa-large}     &  51.1 &	62.6 &	56.8	&  35.4 &	51.8	& 43.6 &	49.7 &	64.7 &	57.2 \\
FT \#512   \texttt{RoBERTa-large}    &   84.3 &	54.8 &	69.5 &	82.1 &	29.6 &	55.8 &	88.1 &	50.1 &	69.1  \\
\midrule
zero-shot   \texttt{RoBERTa-base}     & 48.2 &	58.1 &	53.15 &	37.3 &	41.5 &	39.4 &	48.8 &	56.4 &	52.6 \\
FT \#512   \texttt{RoBERTa-base}    &   74.4 &	49.9 &	62.15 &	79.0 &	26.9 &	52.9 &	83.7 &	48.5 &	66.1 \\
\midrule
zero-shot   \texttt{BERT-large-uncased} & 45.3 &	55.4 &	50.4 &	34.7 &	33.4 &	34.0 &	41.5 &	54.8 &	48.1 \\
FT \#512   \texttt{BERT-large-uncased} & 70.9 &	50.0 &	60.4 &	77.3 &	26.3 &	51.8 &	79.9 &	49.5 &	64.7 \\
\midrule
zero-shot   \texttt{BERT-base-uncased} & 43.5 &	55.9 &	49.7 &	40.7 &	50.8 &	45.8 &	38.7 &	49.9 &	44.3 \\
FT \#512   \texttt{BERT-base-uncased} & 63.2 &	50.1 &	56.65 &	73.9 &	29.1 &	51.5 &	74.5 &	42.6 &	58.5 \\
\bottomrule        
\end{tabular}
\caption{Evaluation results of different pretrained language models. Models are evaluated against both the in-distribution (\textbf{In.}) set and corresponding challenge set.}
\label{tab:models_results}
\end{table*}

\begin{figure}%
\centering
\includegraphics[height=4.45cm]{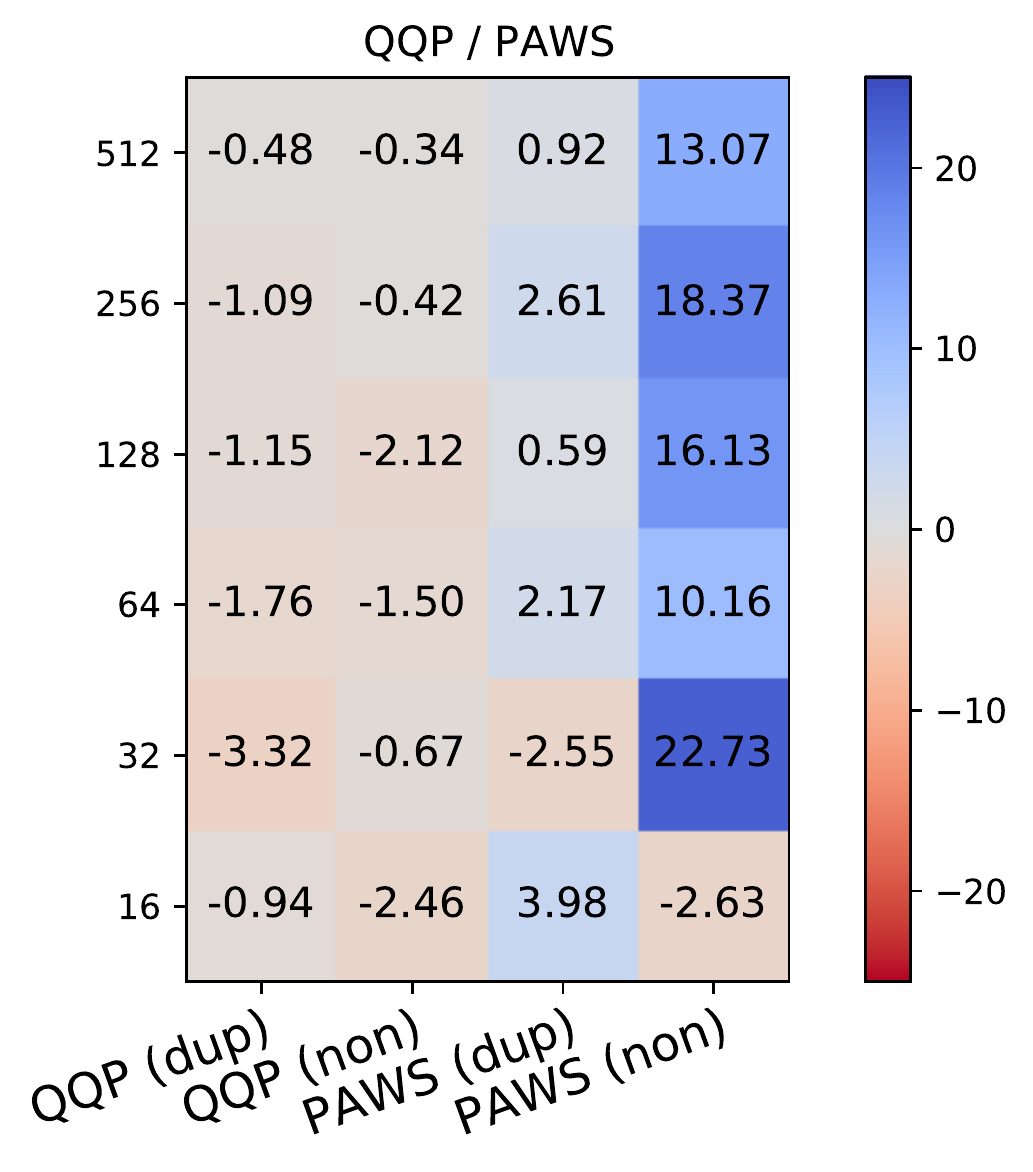}
\includegraphics[height=4.45cm]{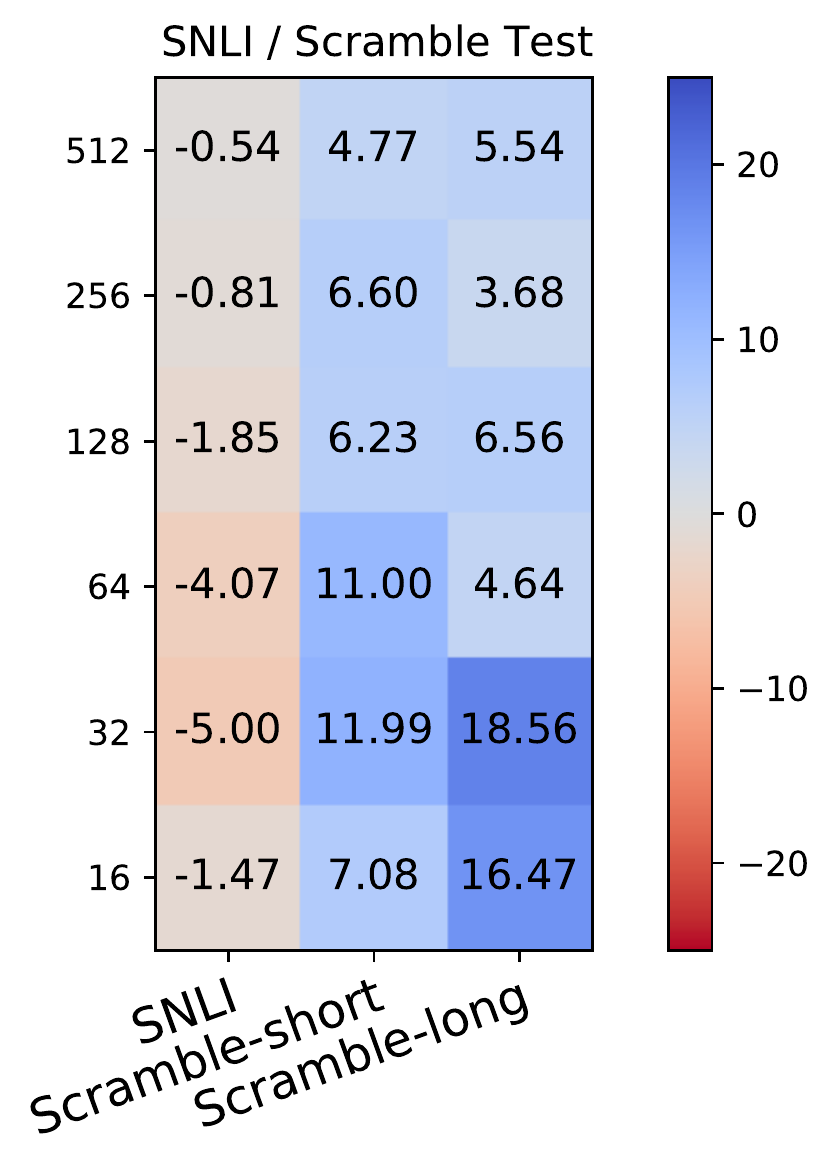}
\caption{Relative difference between median accuracy of prompt-based finetuning across data regimes (y axis) with and without regularization on QQP / PAWS and SNLI / Scramble Test.}
\label{fig:ewc_results2}
\end{figure}

\paragraph{Impact of prompt templates} A growing number of work propose varying prompt generation strategies to push be benefits of prompt-based predictions \cite{gao-etal-2021-making, schick-etal-2020-automatically}. We therefore questions whether different choices of templates would affect the model's behavior related to lexical overlap. We evaluate the 3 top-performing templates for MNLI that are obtained automatically by \citet{gao-etal-2021-making} and show the results in Table \ref{tab:templates_results}. We observe similar behavior from the resulting models over the manual prompt counterpart, achieving HANS average accuracy of around 62\% and below 55\% on zero-shot and finetuning with 512 examples.

\paragraph{Impact of learning steps} We investigate the degradation of the challenge datasets performance as the function of the number of training data available during finetuning. However, adding more training examples while fixing the number of epochs introduces a confound factor to our finding, which is the number of learning steps to the model's weights. To factor out the number of steps, we perform similar evaluation with a fixed amount of training data and varying number of training epochs. On 32 examples per label, we finetune for 10, 20, 30, 40, and 50 epochs. Additionally, we finetune on 512 examples for 1 until 10 epochs to see if the difference in learning steps results in different behavior. We plot the results in Figure \ref{fig:epoch_n}. We observe that both finetuning settings result in similar trajectories, i.e., models start to adopt heuristics immediately in early epochs and later stagnate even with increasing number of learning steps. For instance, finetuning on 32 examples for the same number of training steps as in 512 examples finetuning for 1 epoch still result in higher overall HANS performance. We conclude that the number of finetuning data plays a more significant role over the number of training steps. Intuitively, larger training data is more likely to contain more examples that disproportionately \textit{support} the heuristics; e.g. NLI pairs with lexical overlap are rarely of non-entailment relation \cite{mccoy-etal-2019-right}.

\paragraph{Regularization across data regimes} Figure \ref{fig:ewc_results2} shows the results improvement of L2 weight regularization over vanilla prompt-based finetuning on QQP and SNLI. Similar to results in MNLI/HANS, the improvements are highest on mid data regimes, e.g., 32 examples per label.

\paragraph{Impact of pretrained model} In addition to evaluating \texttt{RoBERTa-large}, we also evaluate on other commonly used pretrained language models based on transformers such as \texttt{RoBERTa-base}, \texttt{BERT-base-uncased}, and \texttt{BERT-large-uncased}. The results are shown in Table \ref{tab:models_results}. We observe similar pattern across PLMs, i.e., improved in-distribution scores come at the cost of the degradation in the corresponding challenge datasets.


\end{document}